\documentclass[10pt,twocolumn,letterpaper]{article}

\usepackage{iccv}
\usepackage{times}
\usepackage{epsfig}
\usepackage{epstopdf}
\usepackage{float}
\usepackage{graphicx}
\usepackage{amsmath}
\usepackage{amssymb}
\usepackage{rotating}
\usepackage{booktabs}
\usepackage{moresize}
\usepackage[ruled]{algorithm2e}

\newcommand{\RR}[2]{\mathbb{R}^{#1 \times #2}}
\newcommand{\Bin}[2]{\{0,1\}^{#1 \times #2}}

\newcommand{\InProd}[2]{\langle #1, #2\rangle}

\newcommand{\trace}[1]{\mbox{trace}\left(#1\right)}
\newcommand{\rank}[1]{\mbox{rank}\left(#1\right)}

\newcommand{\st}{\mbox{s.t.~}}

\newcommand{\refEq}[1]{(\ref{#1})}
\newcommand{\refFig}[1]{Figure~\ref{#1}}

\newcommand{\refSec}[1]{Section~\ref{#1}}

\newcommand{\refAlg}[1]{Algorithm~\ref{#1}}
\newcommand{\refTab}[1]{Table~\ref{#1}}

\def\bfA{{\boldsymbol{A}}}
\def\bfB{{\boldsymbol{B}}}

\def\bfI{{\boldsymbol{I}}}

\def\bfS{{\boldsymbol{S}}}

\def\bfW{{\boldsymbol{W}}}
\def\bfX{{\boldsymbol{X}}}
\def\bfY{{\boldsymbol{Y}}}

\def\bfzero{{\boldsymbol{0}}}
\def\bfone{{\boldsymbol{1}}}

\newcommand{\ProjToC}[1]{\mathcal{P}_{\mathcal{C}}\left(#1\right)}

\usepackage[pagebackref=true,breaklinks=true,letterpaper=true,colorlinks,bookmarks=false]{hyperref}

\iccvfinalcopy 

\def\iccvPaperID{1186} 

\ificcvfinal\fi

\begin{document}

\title{Multi-Image Matching via Fast Alternating Minimization}

\author{Xiaowei Zhou, Menglong Zhu, Kostas Daniilidis \\
GRASP Laboratory, University of Pennsylvania\\
{\tt\small \{xiaowz,menglong,kostas\}@seas.upenn.edu}
}

\maketitle

\begin{abstract}
In this paper we propose a global optimization-based approach to jointly matching a set of images. The estimated correspondences simultaneously maximize pairwise feature affinities and cycle consistency across multiple images. Unlike previous convex methods relying on semidefinite programming, we formulate the problem as a low-rank matrix recovery problem and show that the desired semidefiniteness of a solution can be spontaneously fulfilled. The low-rank formulation enables us to derive a fast alternating minimization algorithm in order to handle practical problems with thousands of features. Both simulation and real experiments demonstrate that the proposed algorithm can achieve a competitive performance with an order of magnitude speedup compared to the state-of-the-art algorithm. In the end, we demonstrate the applicability of the proposed method to match the images of different object instances and as a result the potential to reconstruct category-specific object models from those images.
\end{abstract}

\maketitle

\section{Introduction}

Finding feature correspondences between two images is a fundamental problem in computer vision with various applications such as structure from motion, image registration, shape analysis, to name a few.  While previous efforts were mostly focused on matching a pair of images, many tasks require to find correspondences across multiple images. A typical example is nonrigid structure from motion \cite{bregler2000recovering,dai2012simple}, where one can hardly reconstruct a nonrigid shape from two frames. Furthermore, recent work has shown that leveraging multi-way information can dramatically improve matching results compared to pairwise matching \cite{pachauri2013solving,huang2013consistent}.

\begin{figure}
\centering
\includegraphics[width=0.8\linewidth]{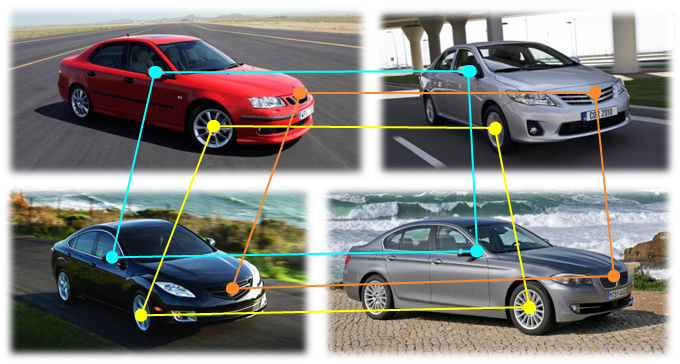}
\caption{An illustration of consistent multi-image matching. } \label{fig:demo}
\vspace{-1.5em}
\end{figure}

The most important constraint for joint matching is the cycle consistency, i.e., the composition of matches along a loop of images should be identity, as illustrated in \refFig{fig:demo}. Given pairwise matches, one can possibly identify true or false matches by checking all cycles in the image collection. But there are many difficulties for this approach \cite{chen2014near}. For example, the input pairwise matches are often very noisy with many false matches and missing matches, and the features detected from different images may only have a partial overlap even if the same feature detector is applied \cite{mikolajczyk2005comparison}. Therefore, it is likely that very few consistent cycles can be found. Moreover, how to sample cycles is not straightforward due to the huge number of possibilities \cite{huang2013consistent}. Recent work on joint matching has shown that, if all feature correspondences within multiple images are denoted by a large binary matrix, the cycle consistency can be translated into the fact that such a matrix should be positive semidefinite and low-rank \cite{kim2012exploring,pachauri2013solving,huang2013consistent}. Based on this observation, convex optimization-based algorithms were proposed, which achieved the state-of-the-art performances with theoretical guarantees \cite{huang2013consistent,chen2014near}. But these algorithms rely on semidefinite programming (SDP), which is not computationally efficient to handle image matching problems in practice.

In this paper, we propose a novel algorithm for multi-image matching. The inputs to our algorithm are original similarities between feature descriptors such as SIFT descriptors \cite{lowe2004distinctive} and deep features \cite{weinzaepfel2013deepflow}, or optimized affinities provided by existing graph matching solvers \cite{leordeanu2005spectral}. The outputs are feature correspondences between all pairs of images. Unlike many previous methods starting from quantized pairwise matches \cite{pachauri2013solving,chen2014near}, we postpone the decision until we optimize for both pairwise affinities and multi-image consistency. Instead of using SDP relaxation, we formulate the problem as a low-rank matrix recovery problem and employ the nuclear-norm relaxation for rank minimization (\refSec{sec:formulation}). We show that the positive semidefiniteness of a desired solution can be spontaneously fulfilled (\refSec{sec:psd}). Moreover, we derive a fast alternating minimization algorithm to globally solve the problem in the low-dimensional variable space (\refSec{sec:optimization}). Besides validating our method on both simulated and real benchmark datasets, we also demonstrate the applicability of the proposed method combined with deep learning and graph matching to match images with different objects and reconstruct category-specific object models (\refSec{sec:experiments}).

\section{Related work}

The early work on joint matching aimed to select cycle-consistent matches and identify incorrect matches from bad cycles \cite{zach2010disambiguating,nguyen2011optimization}. The assumption for this family of methods is that correct matches are dominant in the raw input. Otherwise, it is difficult to find a sufficient number of closed cycles \cite{huang2013consistent}. Some works proposed to use the cycle consistency as an explicit constraint for sparse feature matching \cite{yan2013joint,yan2014graduated,yan2015consistency,yan2015multi} or pixel-wise flow computation \cite{zhou2015flowweb}, but the resulting optimization problems are nonconvex and can hardly be solved globally. Recent results in \cite{kim2012exploring,huang2013consistent,pachauri2013solving} showed that the consistent matches could be extracted from the spectrum (top eigenvectors) of the matrix composed of all pairwise matches. The rationale behind this spectral technique is that the problem can be formulated as a quadratic integer program and relaxed into a generalized Rayleigh problem. But the relaxation assumes full feature correspondences (bijection) between images \cite{pachauri2013solving}. Recently, Huang and Guibas \cite{huang2013consistent} proposed an elegant solution based on convex relaxation and derived the theoretical conditions for exact recovery. The result is further improved in \cite{chen2014near} by assuming that the underlying rank of the variable matrix can be reliably estimated. In these works, the problem is formulated as SDP, which has a limited computational efficiency in real applications.

Regarding methodology, our work is inspired by the recent advances on low-rank matrix recovery which make use of convex relaxation \cite{candes2009exact,candes2011robust} and explore the underlying low-rank structure to accelerate computation \cite{cabral2013unifying,hastie2014matrix}. Our work is also related to some other problems that aim to find global estimates from pairwise estimates such as rotation averaging \cite{hartley2013rotation,wang2013exact} and model fusion \cite{ye2012robust}.

\section{Preliminaries and notation}

Suppose we have $n$ images and $p_i$ features from each image $i$. The objective is to find feature correspondences between all pairs of images. Before introducing the proposed method, we first give a brief introduction to pairwise matching techniques and the definition of cycle consistency.

\subsection{Pairwise matching}\label{sec:pairwise}

To match an image pair $(i,j)$, one can compute similarities for all pairs of feature points from two images and store them in a matrix $\bfS_{ij}\in\RR{p_i}{p_j}$.

We represent the feature correspondences for image pair $(i,j)$ by a partial permutation matrix $\bfX_{ij}\in\Bin{p_i}{p_j}$, which satisfies the doubly stochastic constraints:
\begin{align}
\bfzero \leq \bfX_{ij}\bfone \leq \bfone, ~~~~, \bfzero \leq \bfX_{ij}^T\bfone \leq \bfone. \label{eq:ds-constr}
\end{align}
To find $\bfX_{ij}$, we can maximize the inner product between $\bfX_{ij}$ and $\bfS_{ij}$ subject to the constraints in \refEq{eq:ds-constr} resulting in a linear assignment problem, which has been well studied and can be efficiently solved by the Hungarian algorithm.

In image matching, spatial rigidity is usually preferred, i.e., the relative location between two features in an image should be similar to that between their correspondences in the other image. This problem is well known as graph matching and formulated as a quadratic assignment problem (QAP). While QAP is NP-hard, many efficient algorithms have been proposed to solve it approximately, e.g.,  \cite{leordeanu2005spectral,berg2005shape,cho2010reweighted}. Those solvers basically relax the binary constraint on the permutation matrix, solve the optimization, and output the confidence of a candidate match being correct. We refer readers to the related literature for details. Here we aim to emphasize that the outputs of graph matching solvers are basically optimized affinity scores of candidate matches, which consider both feature similarity and spatial rigidity. We will use these scores (saved in $\bfS_{ij}$) as our input in some cases.

\subsection{Cycle consistency}

Some recent work proposed to use the cycle consistency as a constraint to match a bunch of images \cite{pachauri2013solving,yan2014graduated,chen2014near}. The cycle consistency can be described by
\begin{align}
\bfX_{ij}=\bfX_{iz}\bfX_{zj}, \label{eq:3-cycle}
\end{align}
for any three images $(i,j,z)$ and can be extended to the case with more images.

The recent results in \cite{huang2013consistent,pachauri2013solving} show that the cycle consistency can be described more concisely by introducing a virtual ``universe" that is defined as the set of unique features that appear in the image collection. Each point in the universe may be observed by several images and the corresponding image points should be matched. In this way, consistent matching should satisfy $\bfX_{ij} = \bfA_i\bfA_j^T$, where $\bfA_i\in\Bin{p_i}{k}$ denotes the map from Image $i$ to the universe, $k$ is the number of points in the universe, and $k\geq p_i$ for all $i$.

Suppose the correspondences for all $m=\sum_{i=1}^{n}p_i$ features in the image collection is denoted by $\bfX\in\Bin{m}{m}$:
\begin{align}
\bfX=\left(\begin{array}{cccc}
\bfX_{11} & \bfX_{12} & \cdots & \bfX_{1n}\\
\bfX_{21} & \bfX_{22} & \cdots & \bfX_{2n}\\
\vdots & \vdots & \ddots & \vdots\\
\bfX_{n1} & \cdots & \cdots & \bfX_{nn}
\end{array}\right),\label{eq:bigX}
\end{align}
and all $\bfA_i$s are concatenated as rows in a matrix $\bfA\in\Bin{m}{k}$. Then, one can write $\bfX$ as
\begin{align}
\bfX=\bfA\bfA^T, \label{eq:univ}
\end{align}
From \refEq{eq:univ}, it is clear to see that a desired $\bfX$ should be both positive semidefinite and low-rank:
\begin{align}
\bfX \succeq 0, ~~~~ \rank{\bfX}\leq k. \label{eq:lrsdp-constr}
\end{align}

Using \refEq{eq:lrsdp-constr} the cycle consistency can be effectively imposed without checking all cycles of pairwise matches. Moreover, partial matching is allowed, while bijection needs to be assumed in \refEq{eq:3-cycle}.

\section{Joint matching via rank minimization}

Given affinity scores $\{\bfS_{ij}~|~1\leq i,j \leq n\}$, we aim to find globally consistent matches $\bfX$. Note that $\bfS_{ij}$ can be all-zero if matching is not performed for a pair $(i,j)$. Moreover, affinity scores can be computed from either feature similarities or graph matching solvers according to specific scenarios, as described in \refSec{sec:pairwise}.

\subsection{Formulation} \label{sec:formulation}

We formulate the problem as a low-rank matrix recovery problem. We maximize the inner product between $\bfX_{ij}$ and $\bfS_{ij}$ for all $i$ and $j$ as multiple linear assignment problems. At the same time, we minimize the rank of $\bfX$ to enforce the cycle consistency. We ignore the positive semidefinite constraint on $\bfX$ and will explain the reasons later.

To make the optimization tractable, we make the following relaxations: (1) $\bfX$ is treated as a real matrix $\bfX\in[0,1]^{m\times m}$ instead of a binary matrix, which is a general practice in solving matching problems. Experimentally, we found that the solution values were very close to 0 or 1 and could be stably quantized by a threshold of 0.5. This might be attributed to the existence of a linear term in the cost function \cite{maciel2003global}. (2) Rank of $\bfX$ is replaced by the nuclear norm $\|\bfX\|_*$ (sum of singular values), which is a tight convex relaxation proven to be very effective in various low-rank problems such as matrix completion \cite{candes2009exact} and robust principal component analysis \cite{candes2011robust}.

The estimated $\bfX$ should be sparse since at most one value in each row of $\bfX_{ij}$ can be nonzero. To induce sparsity, we minimize the sum of values in $\bfX$. Combining all three terms, we obtain the following cost function:
\begin{align}\label{eq:raw}
f(\bfX) &= -\sum_{i=1}^{n}\sum_{j=1}^{n} \InProd{\bfS_{ij}}{\bfX_{ij}} + \alpha\InProd{\bfone}{\bfX} + \lambda\|\bfX\|_*, \nonumber \\
&= -\InProd{\bfS-\alpha\bfone}{\bfX} + \lambda\|\bfX\|_*,
\end{align}
where $\InProd{\cdot}{\cdot}$ denotes the inner product and $\bfS\in\RR{m}{m}$ is the matrix collecting all $\bfS_{ij}$s. $\alpha$ is the weight of sparsity, which can be interpreted as a threshold to remove small scores in $\bfS_{ij}$s. In our implementation, we normalize the scores to let them lie between 0 and 1 and empirically set $\alpha=0.1$. $\lambda$ controls the weight of the nuclear norm. We will discuss $\lambda$ in \refSec{sec:psd} and \refSec{sec:param}.

Besides the doubly stochastic constraints in \refEq{eq:ds-constr}, additional constraints shall be imposed on $\bfX$ after relaxation:
\begin{align}
& \bfX_{ii} = \bfI_{p_i}, ~~~~ 1 \leq i \leq n, \label{eq:diag-constr} \\
& \bfX_{ij} = \bfX_{ji}^T, ~~ 1 \leq i,j \leq n, i\neq j, \label{eq:sym-constr} \\
& \bfzero \leq \bfX \leq \bfone, \label{eq:bound-constr}
\end{align}
where \refEq{eq:diag-constr} constrains self-matching to be identity, \refEq{eq:sym-constr} constrains $\bfX$ to be symmetric, and \refEq{eq:bound-constr} constrains the values in $\bfX$ to lie in $[0,1]$.

Finally, we obtain the following optimization problem:
\begin{align}\label{eq:basic}
\min_{\bfX} ~ & \InProd{\bfW}{\bfX} + \lambda \|\bfX\|_*, \nonumber \\
\st ~~ & \bfX \in \mathcal{C},
\end{align}
where $\bfW=\alpha\bfone-\bfS$ and $\mathcal{C}$ denotes the set of matrices satisfying the constraints given in \refEq{eq:ds-constr}, \refEq{eq:diag-constr}, \refEq{eq:sym-constr} and \refEq{eq:bound-constr}.

Upon our experimental observation, the result doesn't degrade noticeably when removing the doubly stochastic constraints in \refEq{eq:ds-constr}. This might be attributed to the existence of the sparsity regularizer. Therefore, we remove \refEq{eq:ds-constr} in implementation to accelerate the computation.

\subsection{Positive semidefiniteness} \label{sec:psd}

We ignore the positive semidefinite constraint for two reasons: (1) solving SDP is generally unscalable; (2) with the constraints in \refEq{eq:diag-constr} and \refEq{eq:sym-constr}, the solution to \refEq{eq:basic} turns out to be nearly positive semidefinite if $\lambda$ is sufficiently large.\footnote{We use the term ``nearly positive semidefinite" to refer to the property that the negative eigenvalues of a matrix, if there exist, are negligible compared to the norm of the matrix.}

Suppose $\sigma_1,\cdots,\sigma_m$ are eigenvalues of $\bfX$. From \refEq{eq:diag-constr}, we have $X_{ii}=1$ for all $i$, and $\sum_{i=1}^{m}\sigma_i=\trace{\bfX}=m$, which implies that the sum of $\sigma_i$s is fixed. From \refEq{eq:sym-constr}, we have $\bfX$ is symmetric, and $\sigma_i$s are all real numbers. When we choose a large $\lambda$, $\|\bfX\|_*=\sum_{i=1}^{m}|\sigma_i|$ dominates the cost function, and a solution with all nonnegative $\sigma_i$s will give the lowest cost, because $\sum_{i=1}^{m}|\sigma_i|\geq\sum_{i=1}^{m}\sigma_i=m$ and the equality holds iff. $\sigma_i\geq 0$ for all $i$.

The boundness $\|\bfX\|_*\geq m$ also implies that the solution to \refEq{eq:basic} will be insensitive to $\lambda$ when $\lambda$ is sufficiently large, and then minimizing the nuclear norm is equivalent to adding a positive semidefinite constraint. The effect of $\lambda$ is experimentally illustrated in \refSec{sec:param}.

\section{Fast alternating minimization}\label{sec:optimization}

\subsection{Optimization in the low-rank space}\label{sec:factorization}

The nuclear norm minimization in \refEq{eq:basic} is convex and the state-of-the-art methods to solve this family of problems are the proximal method \cite{parikh2013proximal} or ADMM \cite{boyd2010distributed} based on iterative singular value thresholding \cite{cai2010singular}. However, singular value decomposition (SVD) needs to be performed in each iteration, which is extremely expensive even for a medium-sized problem. For instance, if there are 20 images with 500 features per image to match, we have to optimize for an $10,000\times 10,000$ matrix. A single SVD for such a matrix takes hundreds of seconds on a typical PC even if a partial SVD solver \cite{larsen2004propack} is used. See \refTab{tab:time-simu} and \refSec{sec:complexity}.

\begin{table}\small
\renewcommand{\arraystretch}{1.3}
\centering
\begin{tabular}{lccccc}
\toprule
    $[n,p]$ & $m=np$ & MatchLift & Partial SVD & MatchALS \\
\hline
$[5,20]$ & $1.0\times10^2$ & 0.005 & 0.016 & 0.001 \\
$[10,20]$ & $2.0\times10^2$ & 0.009 & 0.016 & 0.003 \\
$[20,20]$ & $4.0\times10^2$ & 0.034 & 0.033 & 0.009 \\
$[20,100]$ & $2.0\times10^3$ & 1.472 & 2.023 & 0.283 \\
$[20,500]$ & $1.0\times10^4$ & 166.8 & 219.3 & 9.804 \\
\bottomrule
\end{tabular}
\vspace{0em}
\caption{The CPU time (seconds) for one iteration of MatchALS, MatchLift \cite{chen2014near} and partial SVD \cite{larsen2004propack}. $n$, $p$, and $m$ denote the number of images, the number of points per image, and the dimension of $\bfX$, respectively. We set $k=2p$ for MatchALS. }
\label{tab:time-simu}
\end{table}

Fortunately, recent results on low-rank optimization have shown that one can solve the problem more efficiently via a change of variables $\bfX=\bfA\bfB^T$ \cite{cabral2013unifying,hastie2014matrix}, where $\bfA,\bfB\in\RR{m}{k}$ are new variables with a smaller dimension $k<m$. More importantly, the change of variables will not introduce additional local minima if $k$ is larger than the rank of the original solution. This result was originally derived for low-rank SDP \cite{burer2005local,kulis2007fast} but also applies here since a nuclear-norm minimization problem can be rewritten as SDP \cite{recht2010guaranteed}.

Inspired by these works, we propose the following low-rank factorization-based formulation in order to leverage the underlying low dimensionality of our problem:
\begin{align}\label{eq:als}
\min_{\bfA,\bfB} ~ & \InProd{\bfW}{\bfA\bfB^T} + \lambda \|\bfA\bfB^T\|_*, \nonumber \\
\st ~~ & \bfA\bfB^T \in \mathcal{C}.
\end{align}
Moreover, with the following equation \cite{recht2010guaranteed},
\begin{align}\label{nuclear_fro}
\|\bfX\|_* = \min_{\bfA,\bfB:\bfA\bfB^T=\bfX} ~ \frac{1}{2}\left(\|\bfA\|_F^2+\|\bfB\|_F^2\right),
\end{align}
we finally obtain the following formulation:
\begin{align}\label{eq:als}
\min_{\bfA,\bfB} ~ & \InProd{\bfW}{\bfA\bfB^T} + \frac{\lambda}{2} \|\bfA\|_F^2 + \frac{\lambda}{2} \|\bfB\|_F^2, \nonumber \\
\st ~~ & \bfA\bfB^T \in \mathcal{C}.
\end{align}
The selection of matrix dimension $k$ is critical to the success of change of variables, while it directly affects the computational complexity. We will first provide the algorithm, analyze its complexity and then discuss the selection of $k$.

\subsection{Algorithms}

The problem in \refEq{eq:als} is not straightforward to solve due to the constraint on the product of variables. Instead, we rewrite the problem as
\begin{align}\label{eq:als1}
\min_{\bfX,\bfA,\bfB} ~ & \InProd{\bfW}{\bfX} + \frac{\lambda}{2} \|\bfA\|_F^2 + \frac{\lambda}{2} \|\bfB\|_F^2, \nonumber \\
\st ~~ & \bfX = \bfA\bfB^T, ~~ \bfX \in \mathcal{C},
\end{align}
and apply the ADMM \cite{boyd2010distributed} to solve \refEq{eq:als1}.

The augmented Lagrangian of \refEq{eq:als1} reads:
\begin{align}
    \mathcal{L}_{\mu}\left(\bfX,\bfA,\bfB,\bfY\right) = & \InProd{\bfW}{\bfX} + \frac{\lambda}{2} \|\bfA\|_F^2 + \frac{\lambda}{2} \|\bfB\|_F^2 \\
    + & \InProd{\bfY}{\bfX-\bfA\bfB^T} + \frac{\mu}{2}\|\bfX-\bfA\bfB^T\|_F^2 \nonumber
\end{align}
where $\bfY$ is the dual variable and $\mu$ is a parameter controlling the step size in optimization. We keep the constraint $\bfX\in\mathcal{C}$ since it can be easily handled as we will show later. Then, the ADMM alternately updates each primal variable by minimizing $\mathcal{L}_{\mu}$ and updates the dual variable via gradient ascent while fixing all other variables. The overall algorithm is summarized in \refAlg{alg:MatchALS}.

\begin{algorithm}\label{alg:MatchALS}
\caption{Multi-Image Matching via Alternating Least Squares (MatchALS)}
\LinesNumbered
\KwIn{Pairwise affinity scores $\bfS$}
\KwOut{Globally consistent matches $\bfX$}
randomly initialize $\bfA$ and $\bfB$, $\bfY=\bfzero$ \;
$\bfW=\alpha\bfone-\bfS$ \;
\While{not converged}{
    $\bfA \leftarrow \left(\bfX+\frac{1}{\mu}\bfY\right) \bfB \left(\bfB^T\bfB + \frac{\lambda}{\mu}\bfI\right)^{\dag}$ \;
    $\bfB \leftarrow \left(\bfX+\frac{1}{\mu}\bfY\right) \bfA \left(\bfA^T\bfA + \frac{\lambda}{\mu}\bfI\right)^{\dag}$ \;
    $\bfX \leftarrow \ProjToC{\bfA\bfB^T-\frac{1}{\mu}\left(\bfW+\bfY\right)}$ \;
    $\bfY \leftarrow \bfY^{k} + \mu~\left(\bfX-\bfA\bfB^T\right)$ \;
    }
quantize $\bfX$ with a threshold equal to 0.5.
\end{algorithm}

Minimizing $\mathcal{L}_{\mu}$ over $\bfA$ turns out to be a regularized least squares problem with a closed-form solution given in Step 4 in \refAlg{alg:MatchALS}. The update of $\bfB$ can be solved similarly. The update of $\bfX$ requires to solve:
\begin{align}
\min_{\bfX\in\mathcal{C}} \| \bfX - \bfA\bfB^T + \frac{1}{\mu}\left(\bfW+\bfY\right) \|_F^2,
\end{align}
and the solution turns out to be a projection to $\mathcal{C}$. Since the constraints in $\mathcal{C}$ are all linear, the projection can be solved conveniently. We denote the solution by $\ProjToC{\cdot}$ and leave the details to the supplementary material.

\subsection{Computational complexity}\label{sec:complexity}

The time complexity of an iteration in \refAlg{alg:MatchALS} is dominated by matrix multiplication that requires $O(m^2k)$ flops\footnote{The detail is given in the supplementary material}. We compare it to the state-of-the-art algorithm MatchLift \cite{chen2014near}, which is based on SDP. The time complexity of an iteration in MatchLift is dominated by the eigenvalue decomposition that requires $O(m^3)$ flops. As $m$ is much larger than $k$, MatchALS has a lower complexity compared to MatchLift. Moreover, matrix multiplication is parallelizable and has been inherently multithreaded in Matlab, while the parallelization of eigenvalue decomposition is an open problem. Both MatchALS and MatchLift are based on ADMM and require similar numbers of iterations to converge upon our observation.

The CPU time for some problem sizes is shown in \refTab{tab:time-simu}. The algorithms are implemented in Matlab and tested on a PC with an Intel i7 3.4GHz CPU and 8G RAM. We also compare the time cost of partial SVD using PROPACK \cite{larsen2004propack}, a toolbox widely used to solve large-scale matrix completion problems \cite{lin2010augmented}. In partial SVD, only $r$ leading singular vectors are computed, which is much faster than full SVD when $r/m$ is extremely small. But it is not efficient for a relatively large $r$. In our problem, $r$ should be larger than the true rank and we test partial SVD with $r=p$ in \refTab{tab:time-simu}.

\subsection{Selection of $k$ and rank reduction}\label{sec:rank}

From the previous subsections we see that $k$ determines the complexity of MatchALS and $k$ should be larger than the rank of true solution, i.e. the size of universe. While some spectral techniques have been proposed in previous work for rank estimation \cite{chen2014near}, we found that the estimation was inaccurate when the input was noisy and incomplete. Fortunately, our solution doesn't depend on $k$ if $k$ is larger than the underlying true rank (demonstrated later in \refFig{fig:sensitivity}). A heuristic choice is to set $k=2\hat{r}$, where $\hat{r}$ is a rough estimate of the size of universe.

In real applications, there are likely to be many isolated features in each image which don't have any correspondence in other images. However, the constraint in \refEq{eq:diag-constr} implies that every image feature must be matched to a point in the universe. To see this, recall that we hope $\bfX=\bfA\bfA^T$ in \refEq{eq:univ}. If diagonal values of $\bfX$ are all ones, every row of $\bfA$ has a unit norm, which indicates a match to the universe. Therefore, the size of universe is dramatically increased by those isolated features, and consequently a very large $k$ needs to be selected, which severely increases the computation. To address this issue, we loose the constraint in \refEq{eq:diag-constr} to be
\begin{align}
\trace{\bfX} &= m', \nonumber \\
\mbox{off-diagnal values}\{\bfX_{ii}\} &= \bfzero, ~ 1\leq i \leq n, \label{eq:diag-constr-new}
\end{align}
where $m'\leq m$ is a predefined constant. When $m'=m$, \refEq{eq:diag-constr-new} is reduced to \refEq{eq:diag-constr}. When $m'<m$, we allow some rows and columns in $\bfX$ to be null, which is most likely to happen for the rows and columns corresponding to the isolated features, since ``switching" them off will not lose many affinity scores but be able to reduce the nuclear norm immediately. By using such a ``rank reduction" strategy, the algorithm can automatically prune the isolated features and reduce the size of universe, which enables us to select a smaller $k$ for better computational efficiency. We set $m'=m$ in simulation since there is no isolated feature and $m'=0.7m$ in real experiments.

\section{Experiments}\label{sec:experiments}

\subsection{Simulation}

We evaluate the performance of the proposed method using synthesized data. Given a permutation matrix $\bfX$ and the ground truth $\bfX^*$, we measure the error rate by intersection over union:
\begin{align}
1 - \frac{\left|\tau(\bfX)\cap\tau(\bfX^*)\right|}{\left|\tau(\bfX)\cup\tau(\bfX^*)\right|},
\end{align}
where $\tau$ denotes the matches defined by a permutation matrix and $|\cdot|$ means the size of a set.

\subsubsection{Matching errors}

\begin{figure}
\begin{sideways}
\hspace{2.5em}Spectral
\end{sideways}
\includegraphics[width=0.45\linewidth]{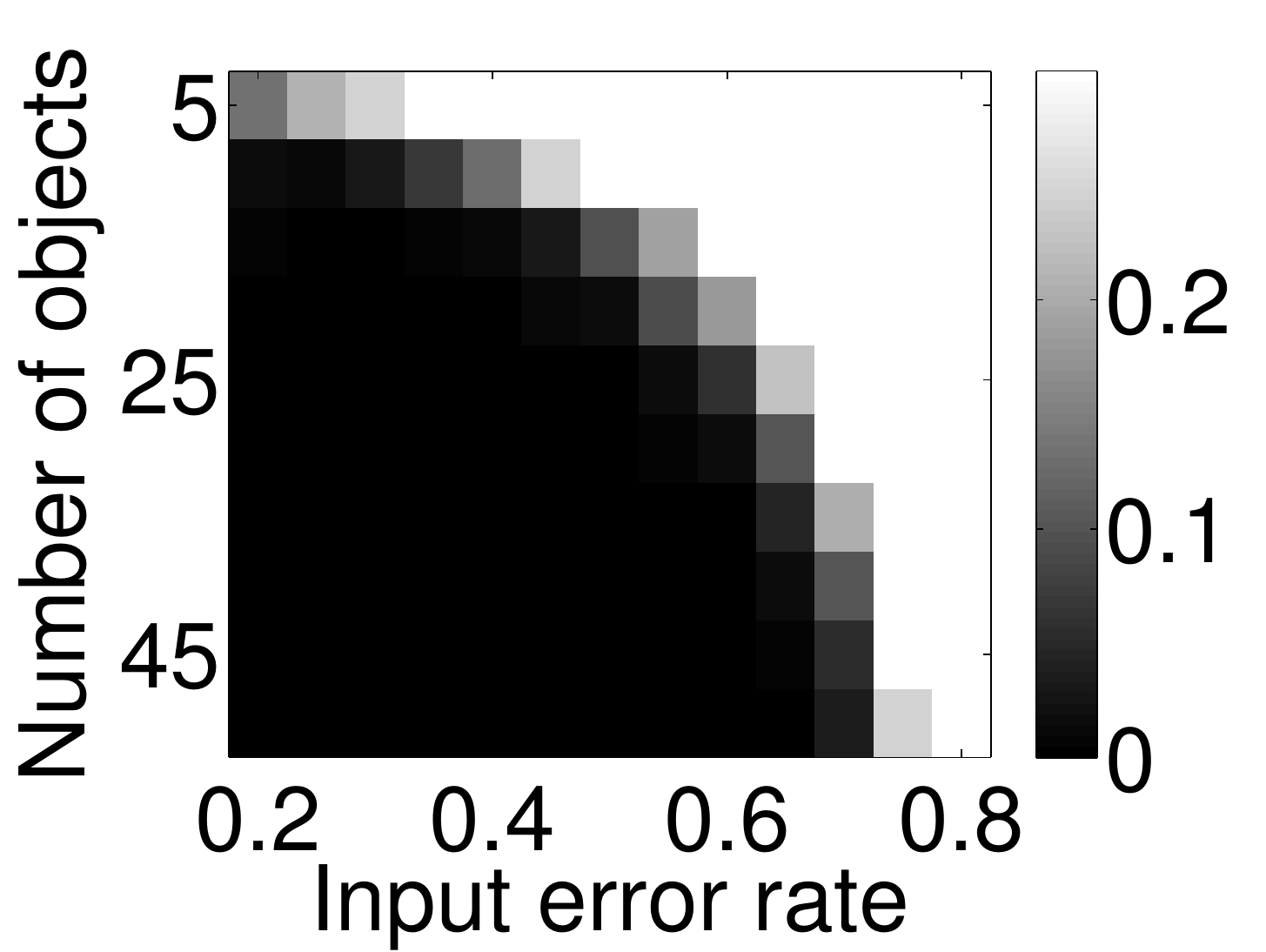}
\includegraphics[width=0.45\linewidth]{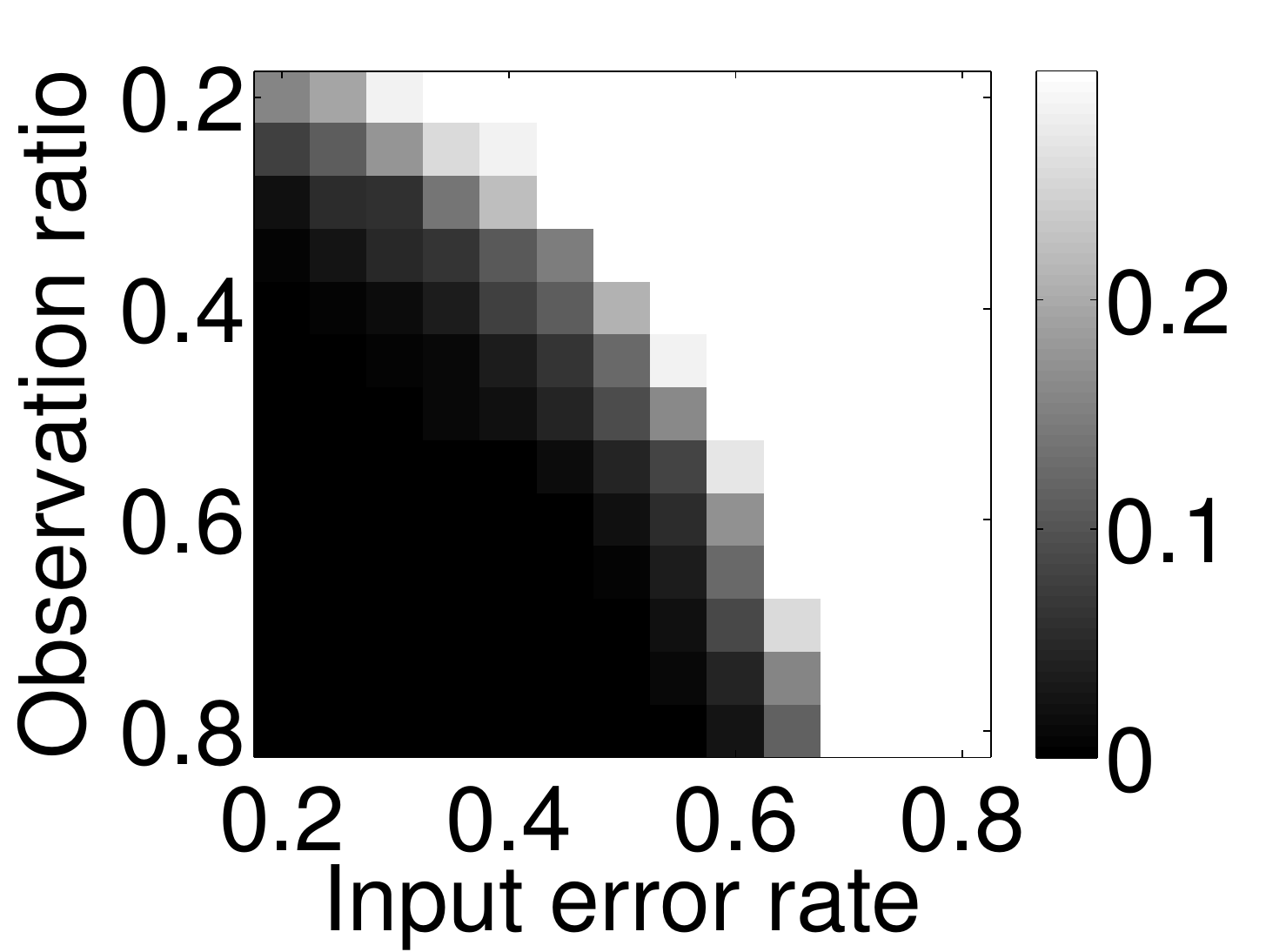}\\
\begin{sideways}
\hspace{2.5em}MatchLift
\end{sideways}
\includegraphics[width=0.45\linewidth]{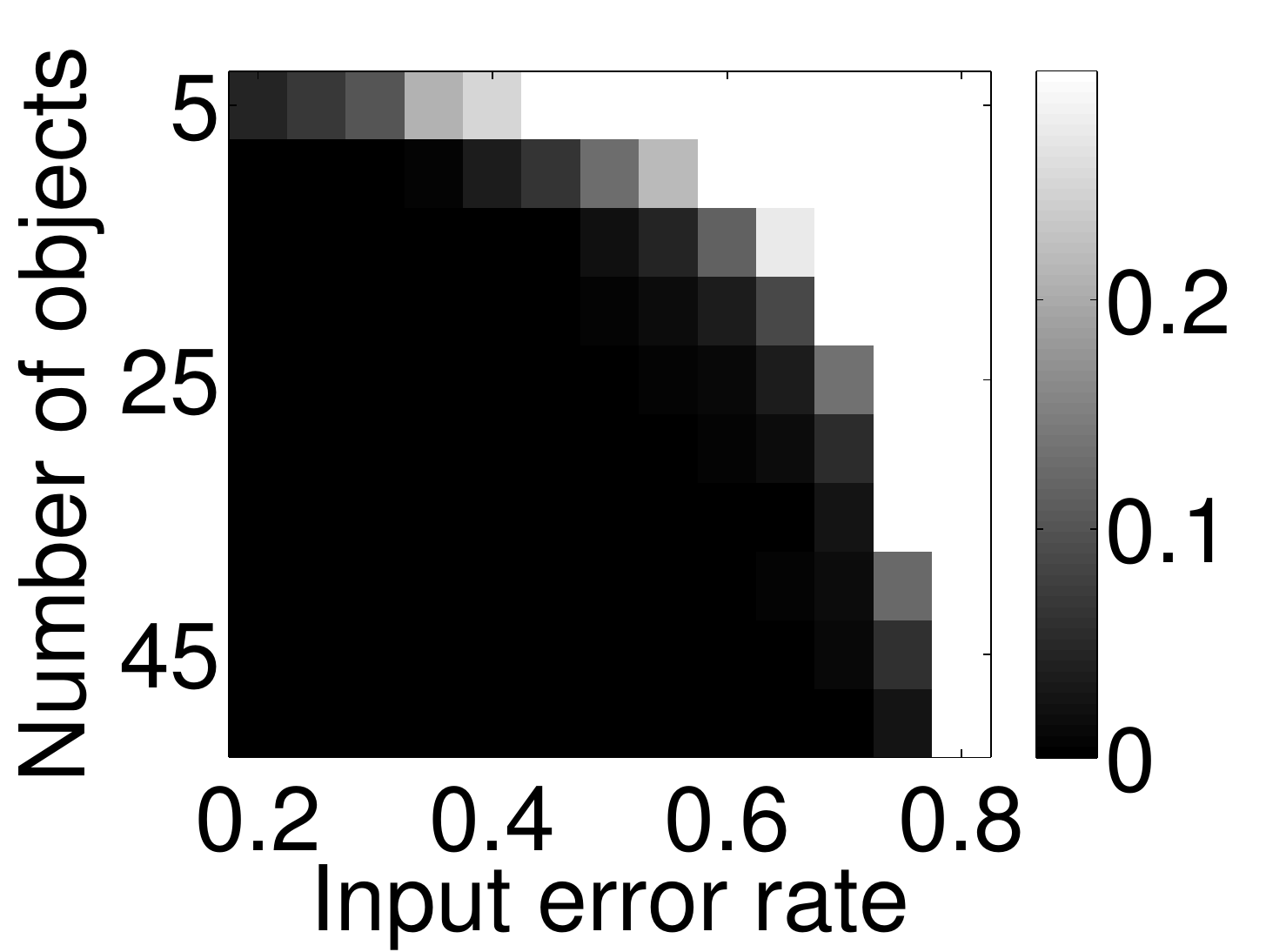}
\includegraphics[width=0.45\linewidth]{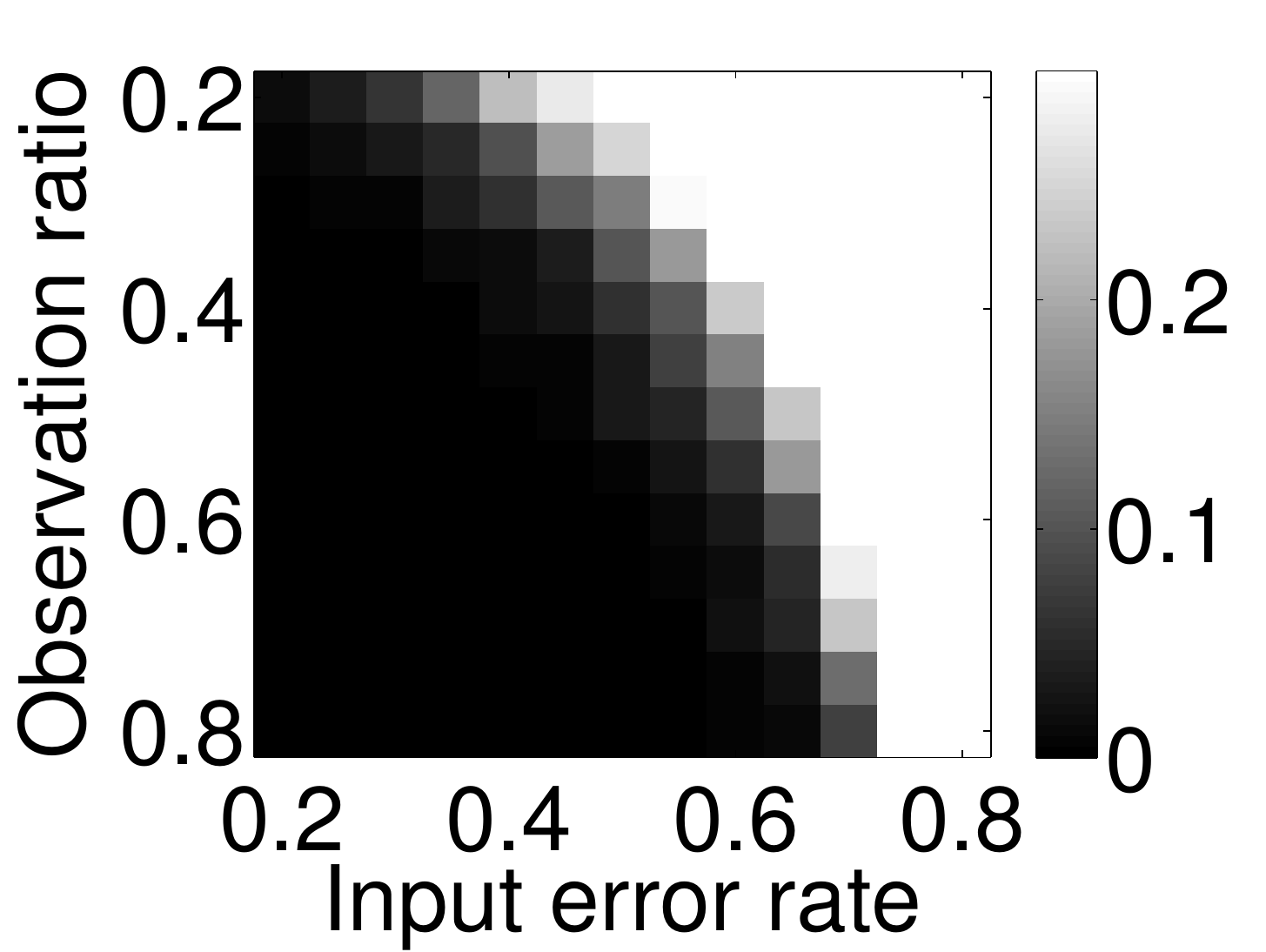}\\
\begin{sideways}
\hspace{2.5em}MatchALS
\end{sideways}
\includegraphics[width=0.45\linewidth]{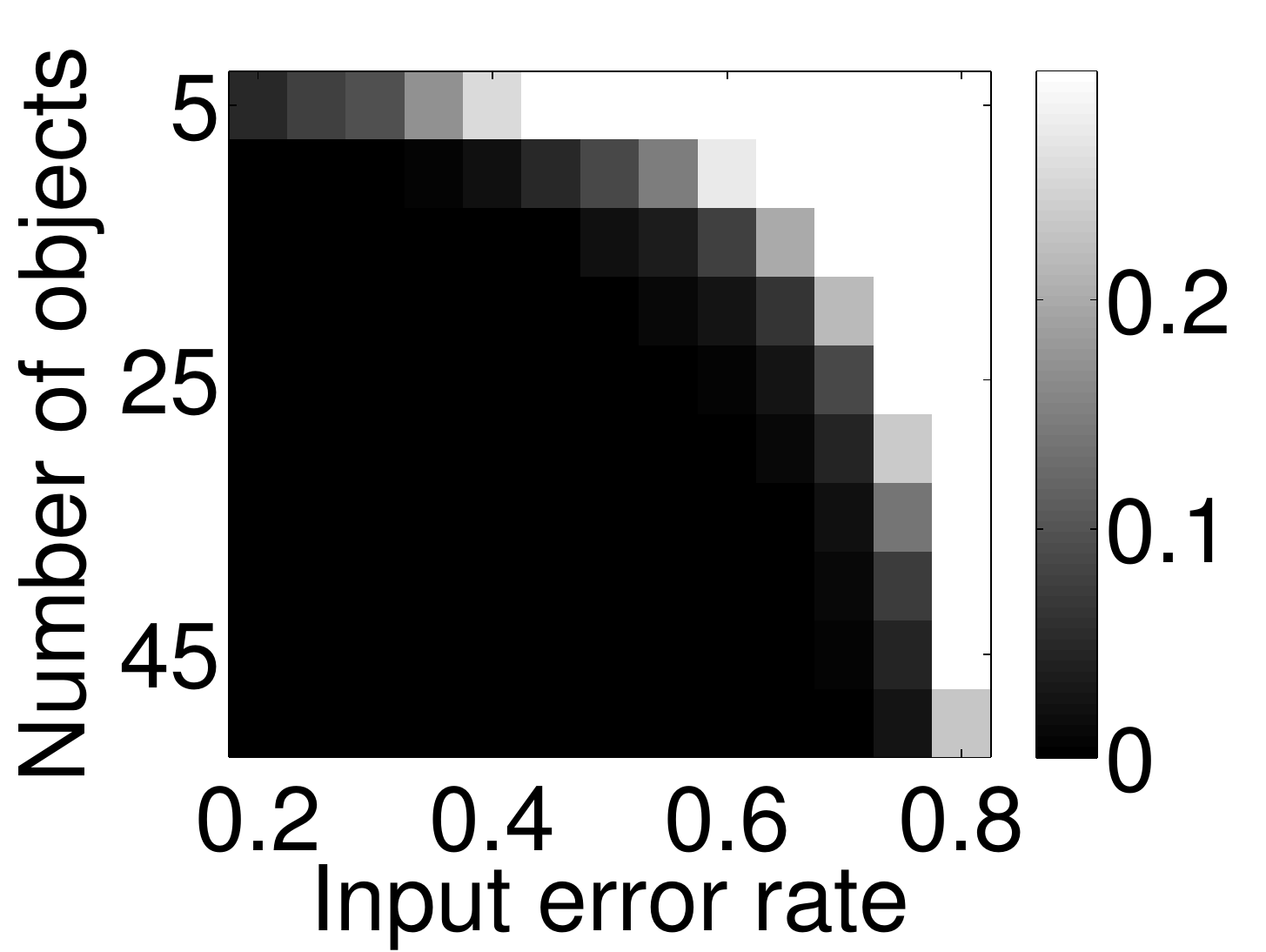}
\includegraphics[width=0.45\linewidth]{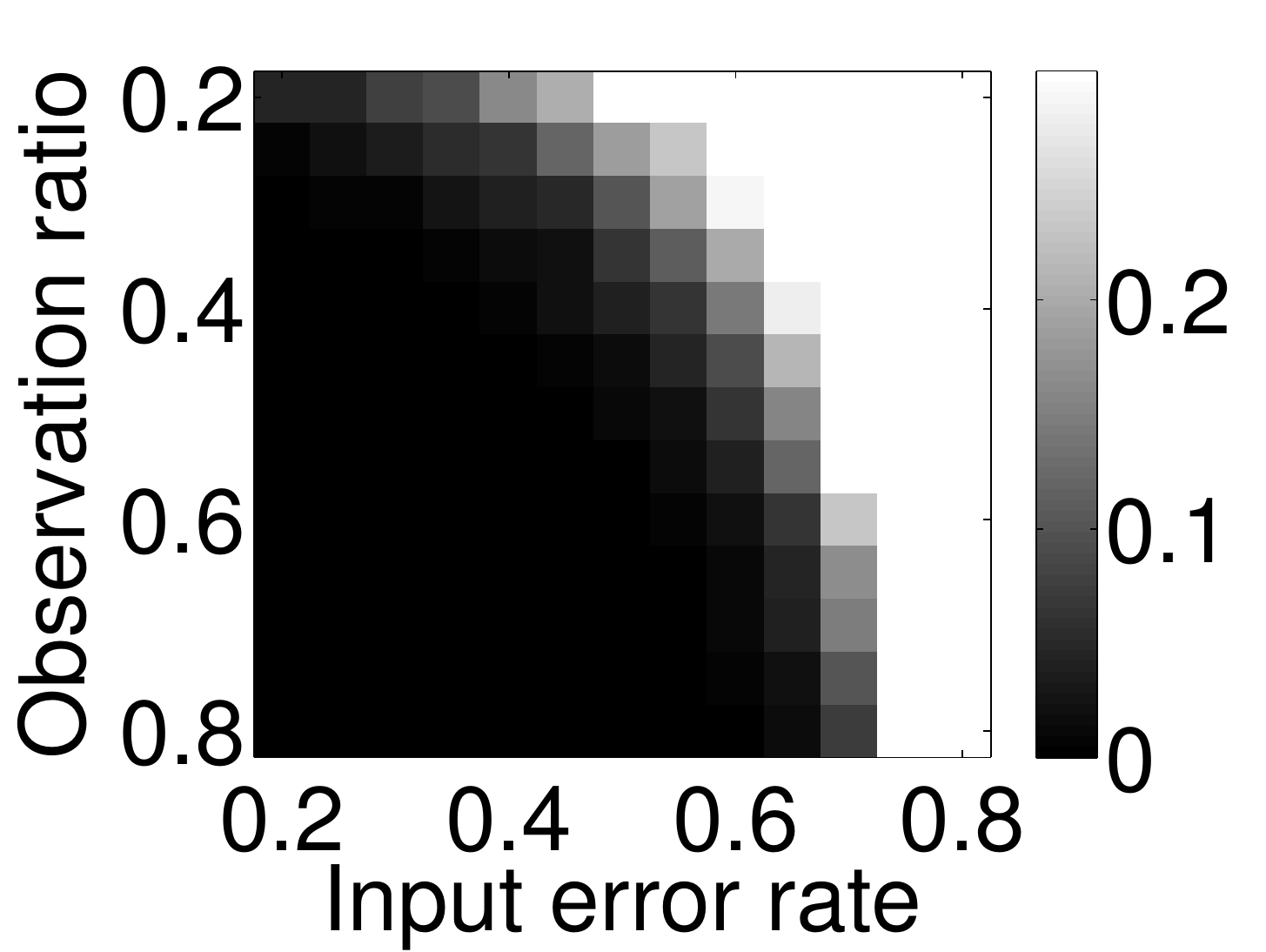}
\caption{The 2D plot of matching errors under various problem settings for the spectral method \cite{pachauri2013solving}, MatchLift \cite{chen2014near} and the proposed MatchALS. In the left column, the number of images $n$ and the input error rate $\rho_e$ are varying, while the observation ratio $\rho_o=0.6$. In the right column, $\rho_o$ and $\rho_e$ are varying, while $n=20$. Lower intensity indicates smaller error and overall a larger dark region indicates a better performance.} \label{fig:simu-error}
\end{figure}

We follow the settings in \cite{chen2014near} to evaluate the performance of MatchALS and compare it to alternative methods. The size of universe is fixed as 20 points and in each image a random sample of the points are observed with a probability denoted by $\rho_o$. The number of images is denoted by $n$. Then, the ground-truth pairwise matches are established, and random corruptions are simulated by removing some true matches and adding some false matches to achieve an error rate of $\rho_e$. Finally, the corrupted permutation matrix is fed into \refAlg{alg:MatchALS} as the input affinity scores.

We evaluate the performance of MatchALS under various $\rho_o$, $\rho_e$ and $n$. We compare MatchALS to two related methods: MatchLift \cite{chen2014near} and the spectral method \cite{pachauri2013solving}. Both of the alternative methods require to know the size of universe and we provide the true value $r^*=20$. For MatchALS parameters, we set $k=2r^*$ and $\lambda=50$.

The output error rates under various settings are shown in \refFig{fig:simu-error}. When the number of images is sufficiently large, all methods can achieve nearly exact recovery even if the input error rate is larger than $50\%$, which demonstrates the power of joint matching. MatchALS and MatchLift achieve very similar performances and outperform the spectral method especially when the observation ratio is small. Compared to MatchLift, the proposed method obtains a competitive performance without exactly knowing the true rank and requires much less computation time.

\subsubsection{Sensitivity to parameters}\label{sec:param}

\begin{figure}
\centering
\includegraphics[width=0.49\linewidth]{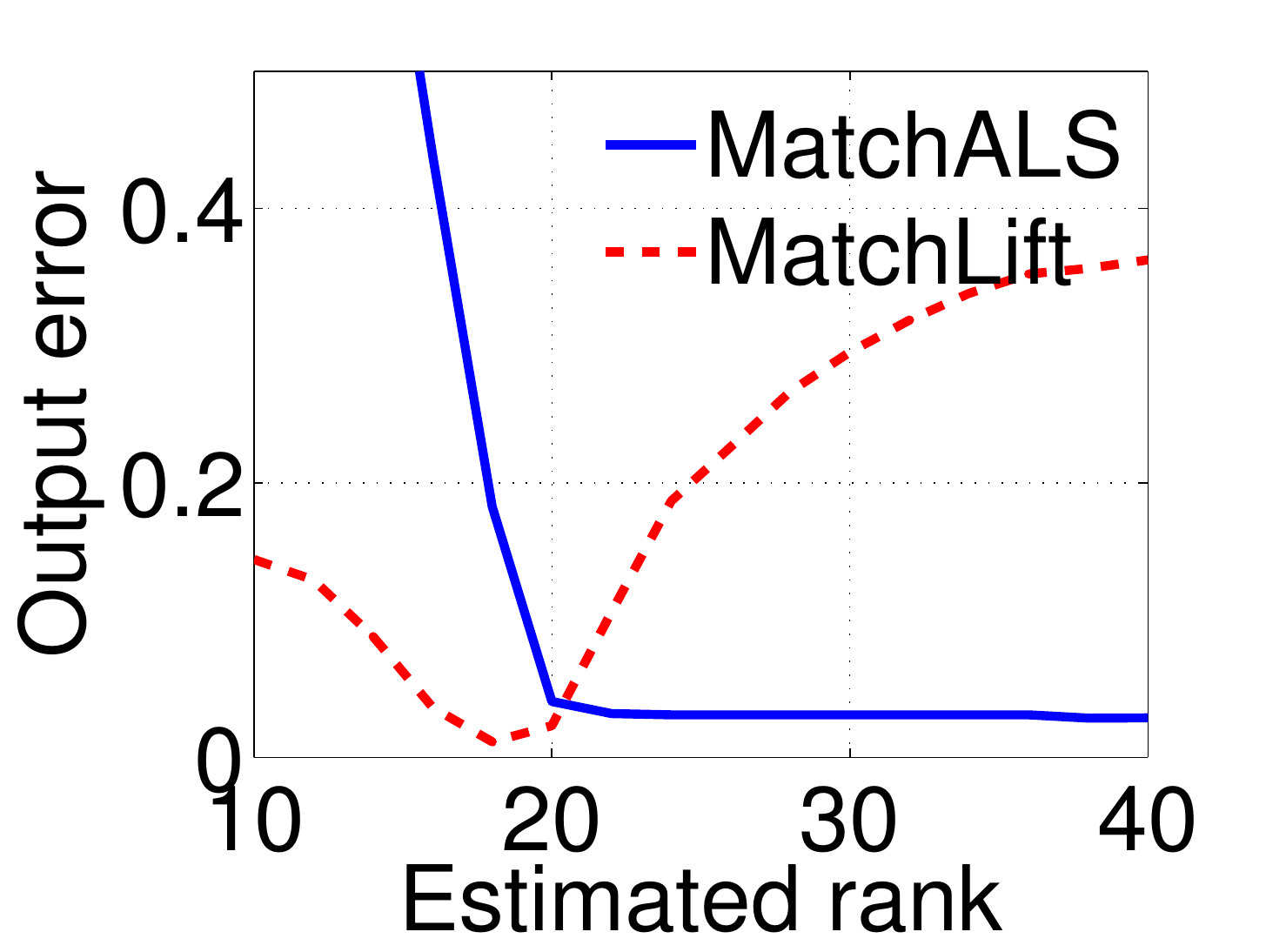}
\includegraphics[width=0.49\linewidth]{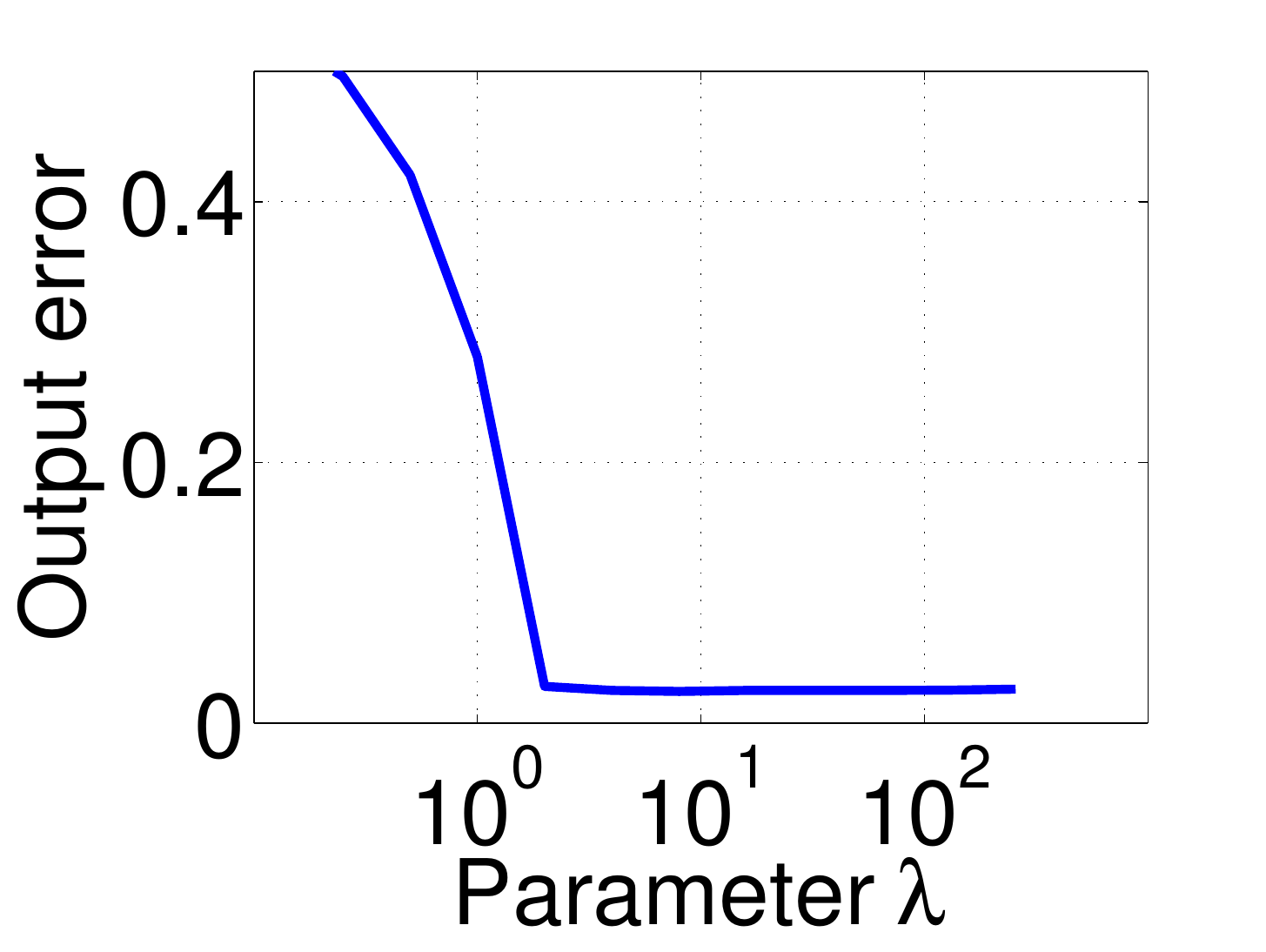}
\caption{The estimation error versus the input rank $\hat{r}$ and the weight of nuclear norm $\lambda$. The true rank $r^*=20$. Here we set $k=\hat{r}$ for MatchALS.} \label{fig:sensitivity}
\vspace{-1em}
\end{figure}

The sensitivity of MatchALS to the parameters in \refEq{eq:als} is illustrated in \refFig{fig:sensitivity}. The figure shows that MatchALS is insensitive to the predefined dimension of factor matrices $k$ when $k$ is larger than the true rank $r^*$, as we explained in \refSec{sec:factorization}. When $k<r^*$, the problem in \refEq{eq:als} is no longer equivalent to the original convex problem in \refEq{eq:basic}, and consequently the alternating minimization fails. In practice, we choose $k=2\hat{r}$ as a compromise between safety and efficiency. The right panel in \refFig{fig:sensitivity} illustrates that the algorithm is insensitive to $\lambda$ when $\lambda$ is sufficiently large as we explained in \refSec{sec:psd}. In all our experiments, we set $\lambda=50$.

\subsection{Real experiments}

\subsubsection{Graffiti datasets}\label{sec:graffiti}

\begin{figure*}
\hspace{7em} Graffiti \hspace{14em} Bikes \hspace{14em} Light \\[-0.1ex]
\includegraphics[width=0.3\linewidth]{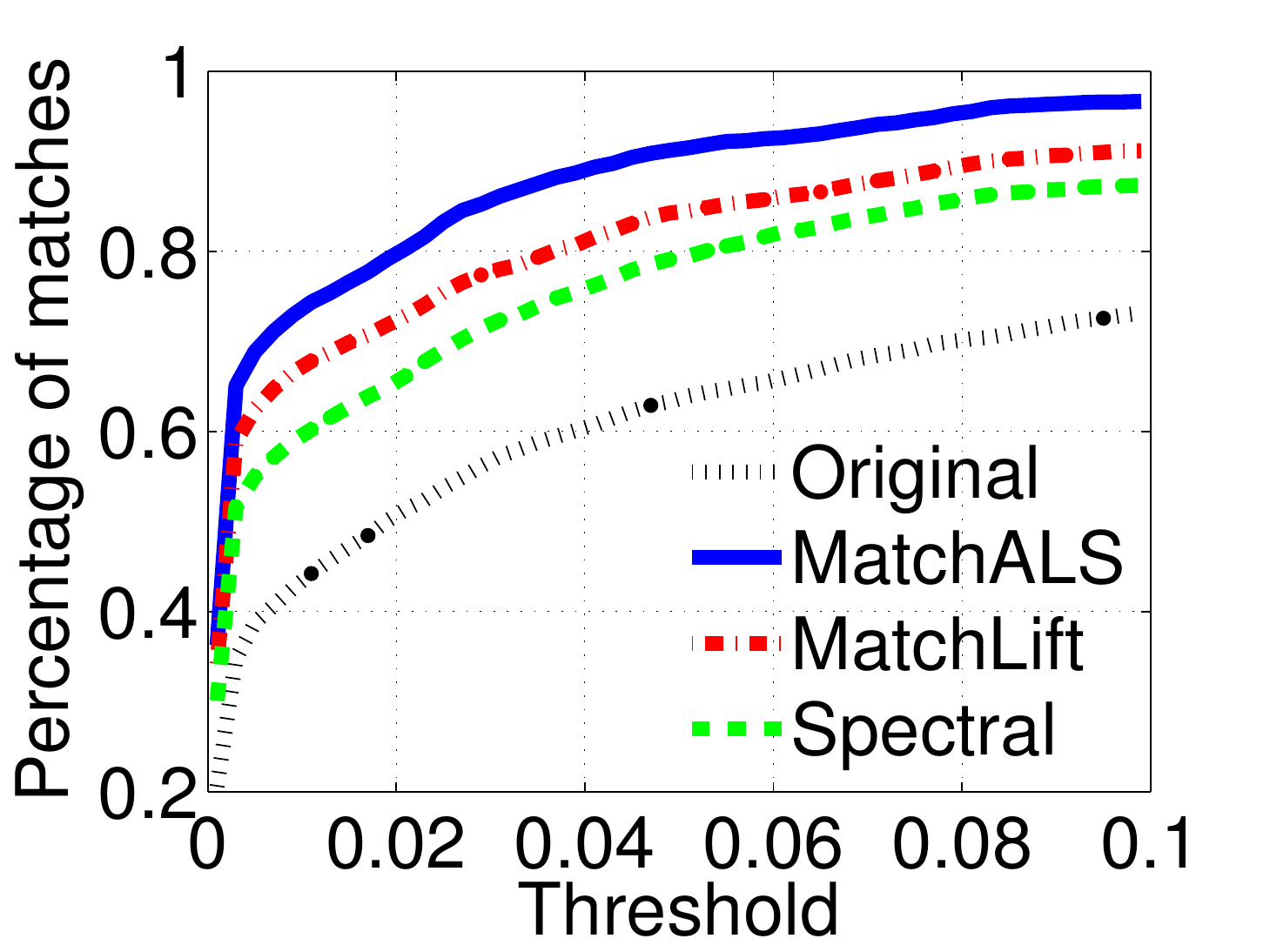} \hfill
\includegraphics[width=0.3\linewidth]{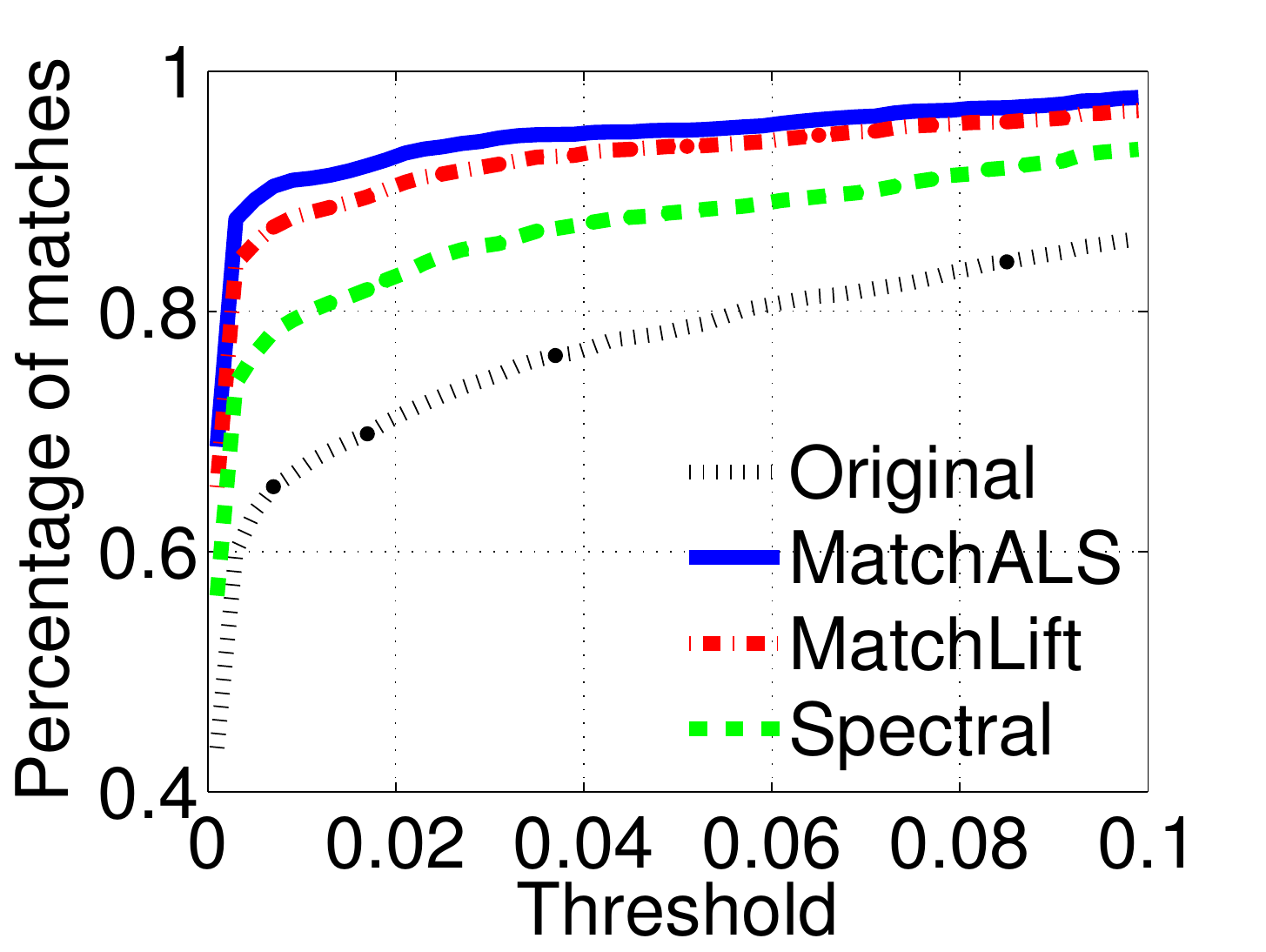} \hfill
\includegraphics[width=0.3\linewidth]{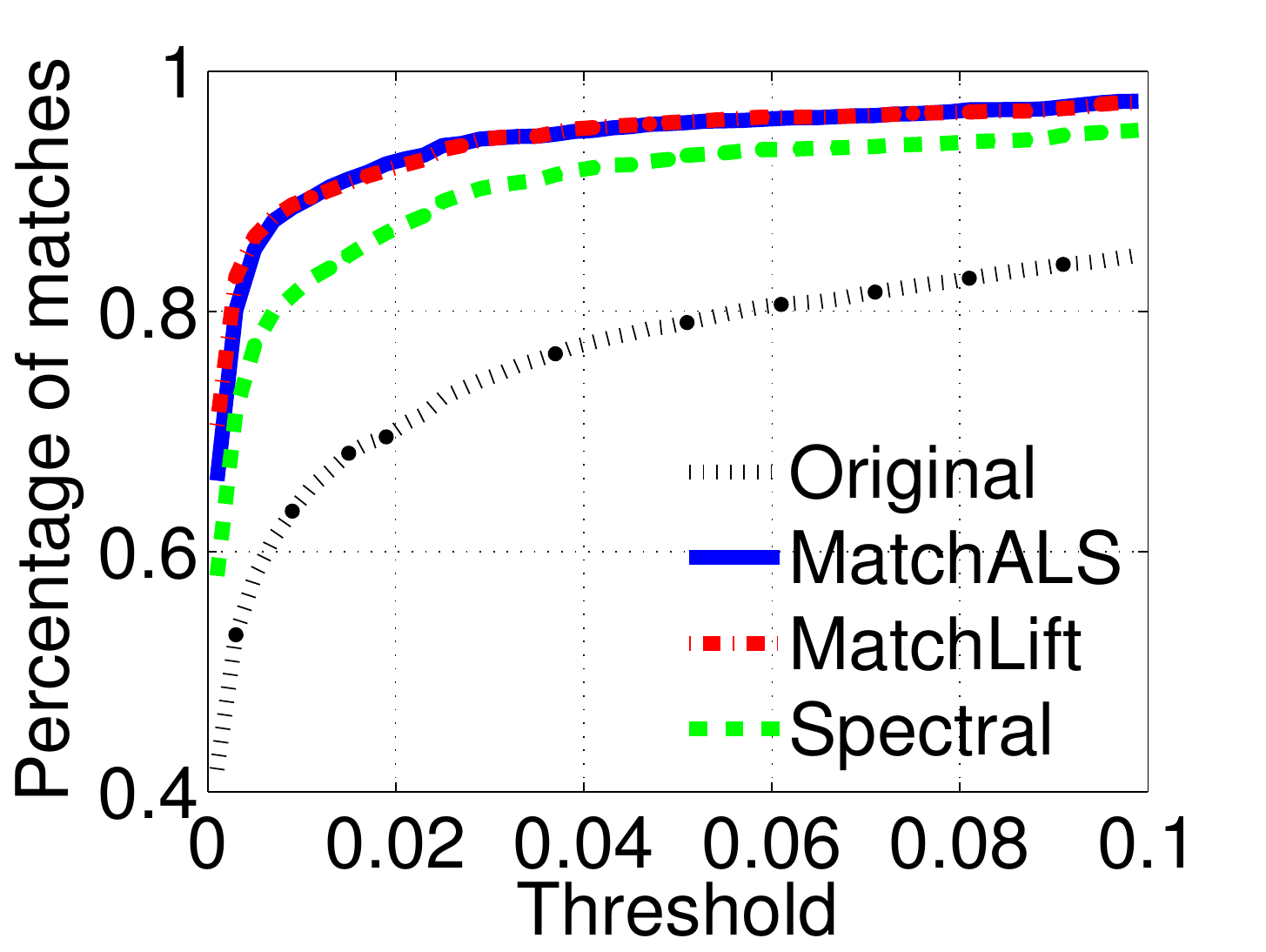}
\caption{The performance curves on the Graff, Bikes and Light datasets. The y-axis shows the percentages of correct matches. The x-axis shows the distance threshold over the image width. Please see \refSec{sec:graffiti} for details. Four methods are compared: MatchALS, MatchLift \cite{chen2014near}, the spectral method \cite{pachauri2013solving}, and the original pairwise matching. The areas under curves for all six datasets are given in \refTab{tab:graffiti}.} \label{fig:graffiti-plot}
\end{figure*}

\begin{figure*}
\centering
\includegraphics[width=0.3\linewidth]{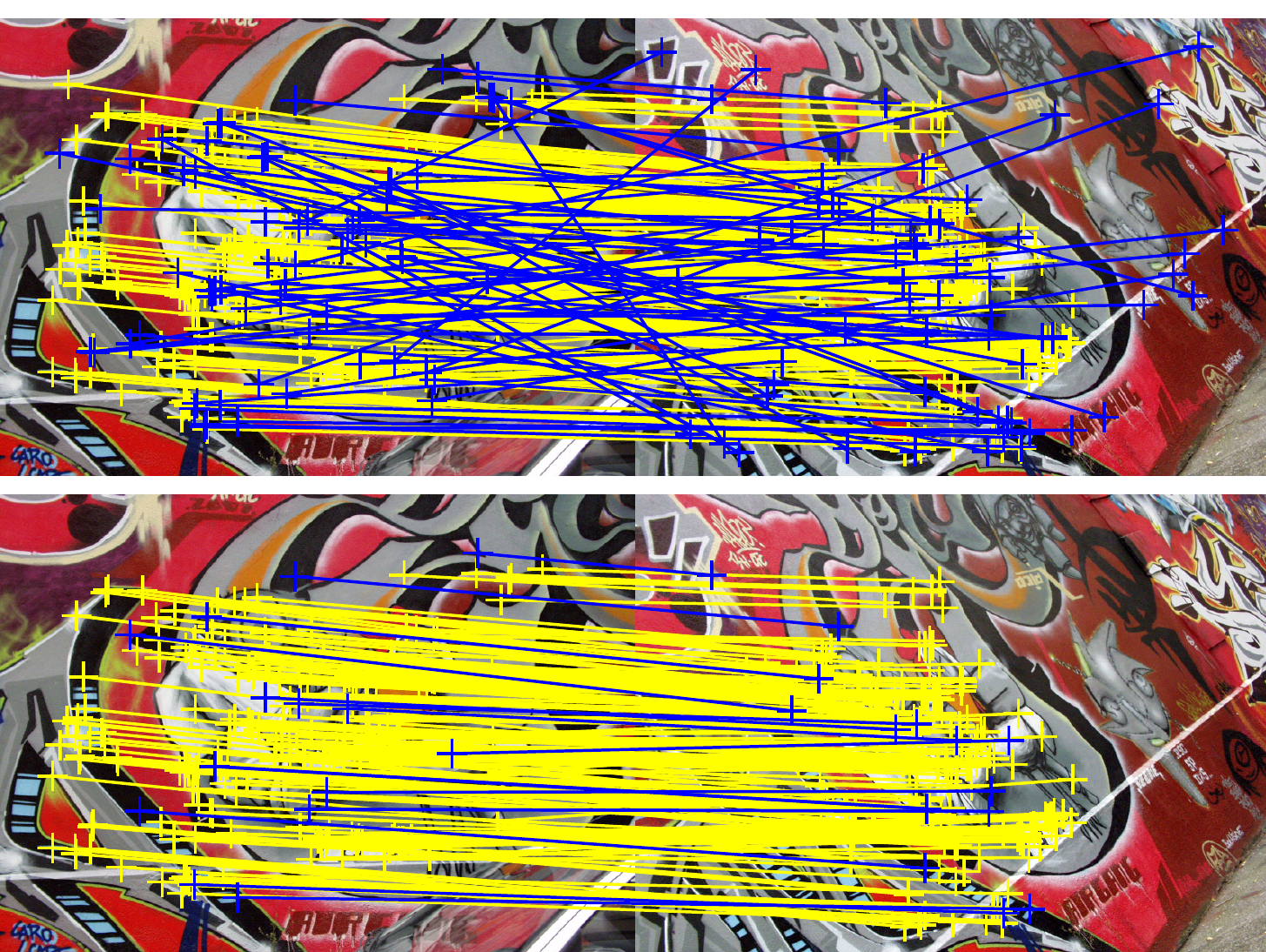} \hfill
\includegraphics[width=0.3\linewidth]{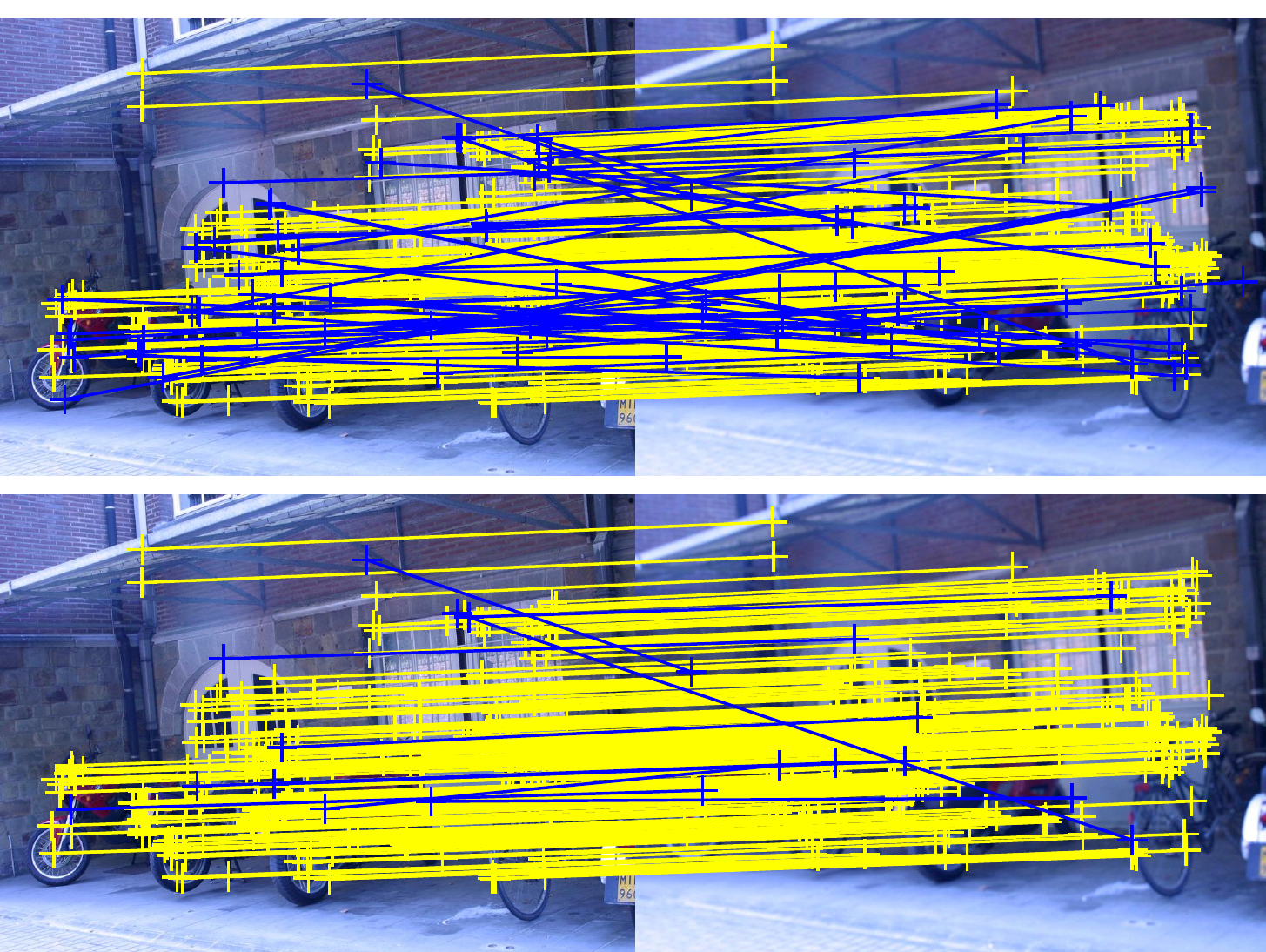} \hfill
\includegraphics[width=0.3\linewidth]{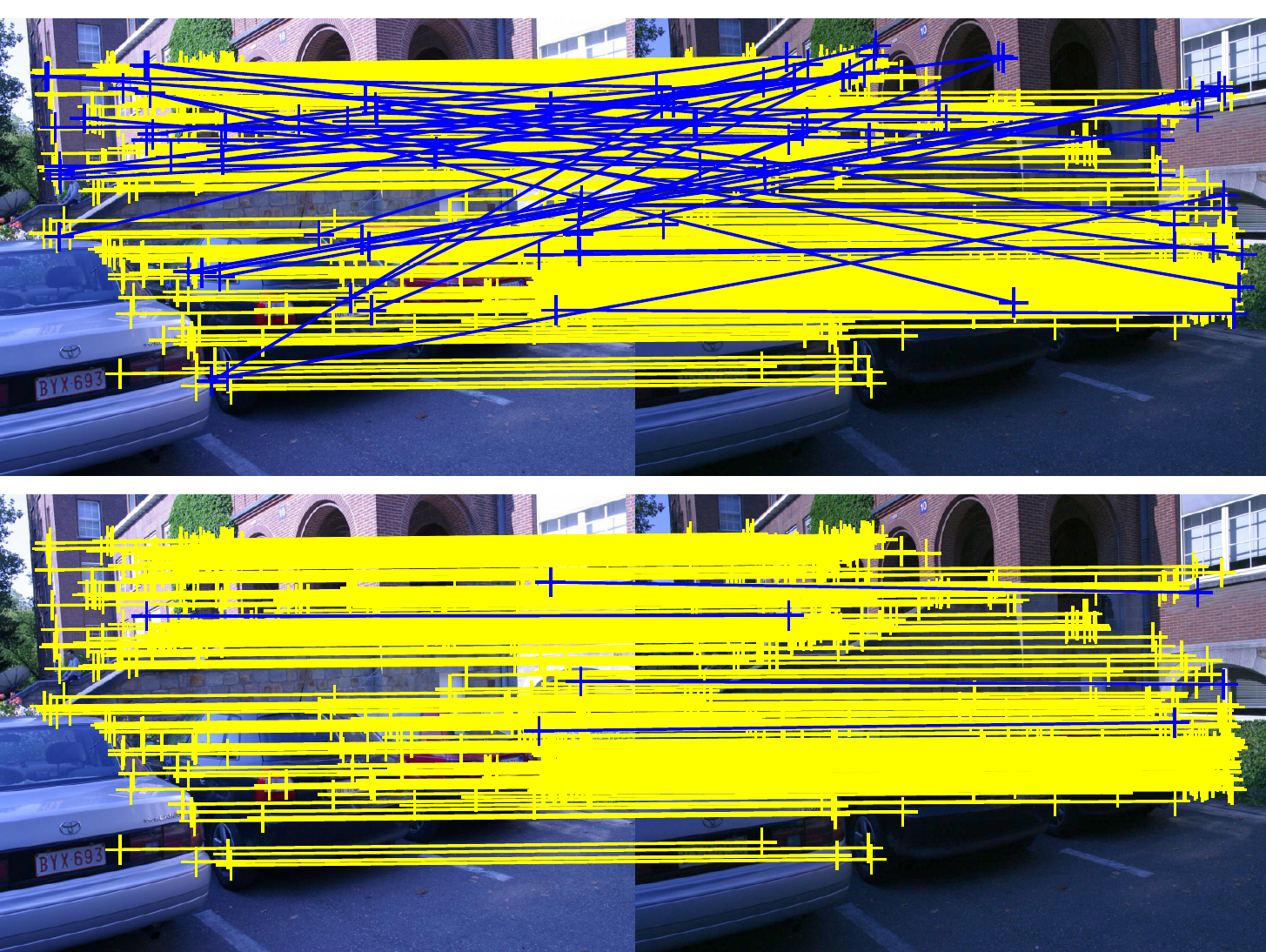}
\caption{The matches between the 1st and the 4th images on the Graff, Bikes and Light datasets. Best viewed in color. The true matches and false matches are shown in yellow and blue, respectively. The top and bottom rows correspond to the results of pairwise matching and joint matching by MatchALS, respectively. } \label{fig:graffiti-example}
\end{figure*}

We evaluate the performance of our algorithm on six benchmark datasets from the Graffiti datasets\footnote{http://www.robots.ox.ac.uk/~vgg/data/data-aff.html}. In each dataset, there are six images of a scene with various image transformations such as viewpoint change, blurring, illumination variation, etc.

We detect 1000 affine covariant features \cite{mikolajczyk2005comparison} with SIFT \cite{lowe2004distinctive} descriptors from each image using the VLFeat library \cite{vedaldi08vlfeat}. For each image pair $(i,j)$, we compute the inner products between feature descriptors as affinity scores and only keep the scores larger than 0.7 and collect them in $S_{ij}$. If the ratio between the first and the second largest scores in a row/column is smaller than 1.1, we set all scores in this row/column to be zero in order to remove indistinctive features. After computing all $\bfS_{ij}$, we remove the features that have candidate matches in less than two images since they have no contribution to joint matching. Finally, we input the affinity scores to \refAlg{alg:MatchALS} to obtain the optimized joint matches.

For evaluation, we adopt the metric used in \cite{chen2014near}: for a testing point in an image, we calculate the distance between its estimated correspondence and the true correspondence in another image. If the distance is smaller than a threshold, we regard that a correct match is found for this testing point. Then, we plot the percentages of testing points with correct matches versus the threshold values and obtain a curve analogous to a precision-recall curve. If a testing point is not aligned with any detected point, its estimated correspondence is obtained by interpolation. In this experiment, we use all detected feature points in the first image as testing points and evaluate the matches from the first image to the other five images. True correspondences are computed from the homography matrices provided in the datasets.

The performance curves on three datasets are shown in \refFig{fig:graffiti-plot}. A curve closer to the upper-left corner indicates a better performance. The area under curve and computation time for all datasets are summarized in \refTab{tab:graffiti}. All of the joint matching methods achieve obvious improvements compared to the original pairwise matching. MatchALS and MatchLift perform similarly and outperforms the spectral method, which coincides with the observation in simulation. Regarding computation time, MatchALS achieves a remarkable speedup ($\sim$30 times on average) compared to MatchLift.

\begin{table}\small
\renewcommand{\arraystretch}{1.3}
\centering
\begin{tabular}{lrrrr}
\toprule
&Original & MatchALS & Spectral & MatchLift \\
\hline
Graffiti & 60.2\% & \bf{87.3\%} & 75.6\% & 80.6\% \\
Bikes & 76.8\% & \bf{94.3\%} & 86.7\% & 92.5\% \\
Boat  & 86.2\% & \bf{93.9\%} & 87.7\% & 91.7\% \\
Light & 76.0\% & 93.9\% & 90.0\% & \bf{94.0\%} \\
Bark  & 71.7\% & \bf{92.2\%} & 91.2\% & 90.0\% \\
UBC   & 88.0\% & \bf{97.0\%} & 92.9\% & 96.8\% \\
\hline
Time  & - & 85.8 & 86.4 & 2518.4\\
\bottomrule
\end{tabular}
\vspace{1em}
\caption{The matching scores and the average computation time (seconds) on the Graffiti datasets. The score is calculated as the area under the curve shown in \refFig{fig:graffiti-plot}.}
\label{tab:graffiti}
\end{table}

We select three image pairs to visually demonstrate the effect of joint matching in \refFig{fig:graffiti-example}. A match with a deviation less than five pixels from the ground truth is declared as true. Clearly, the joint matching can prune the false matches (fewer blue lines), complete some missing matches (denser yellow lines), and achieve almost correct matching for these image pairs with large disparities in viewpoints, blurring and illumination changes.

\subsubsection{Matching different objects}\label{sec:cars}

\begin{figure*}
\centering
\includegraphics[width=0.8\linewidth]{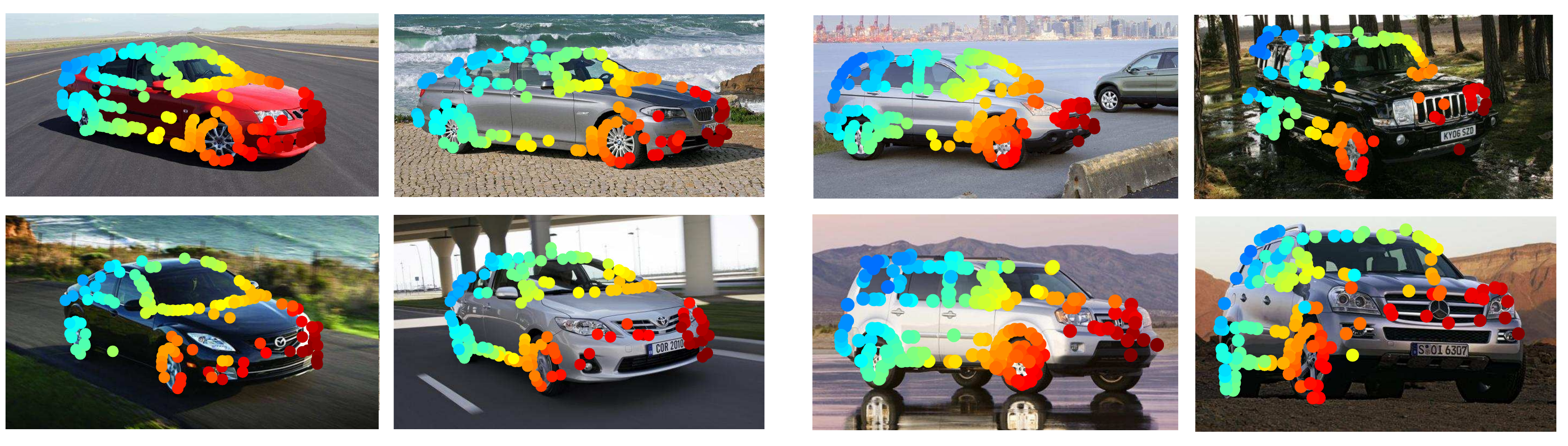}\\
\includegraphics[width=0.4\linewidth]{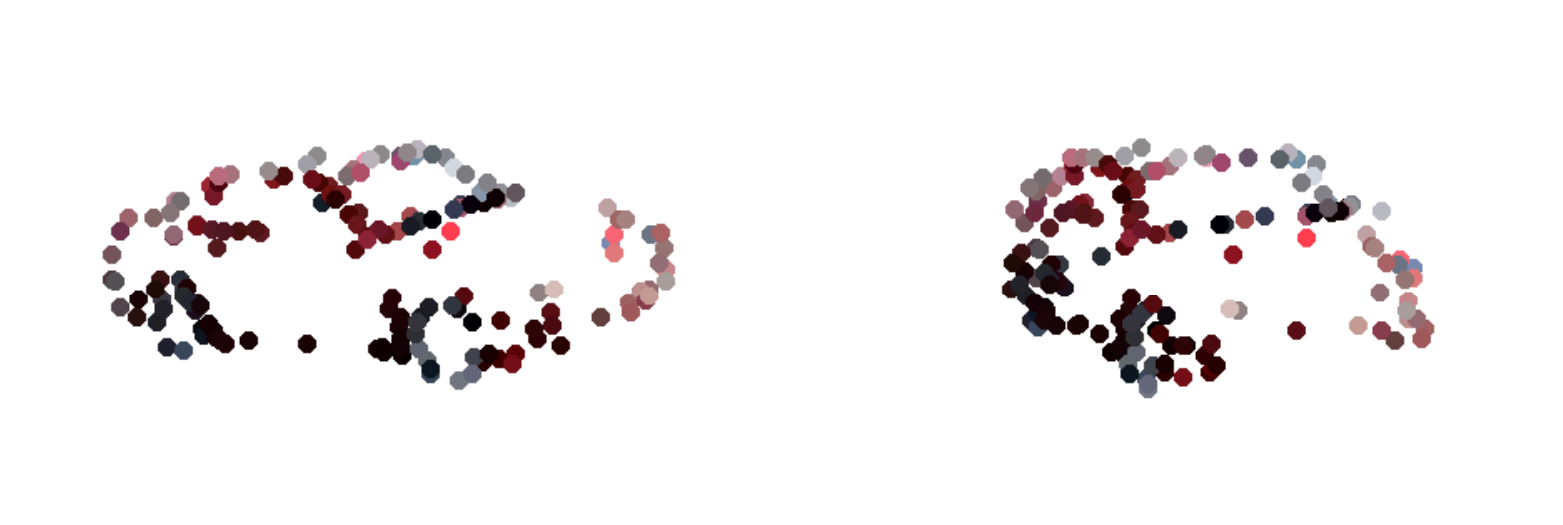} \hspace{1em}
\includegraphics[width=0.4\linewidth]{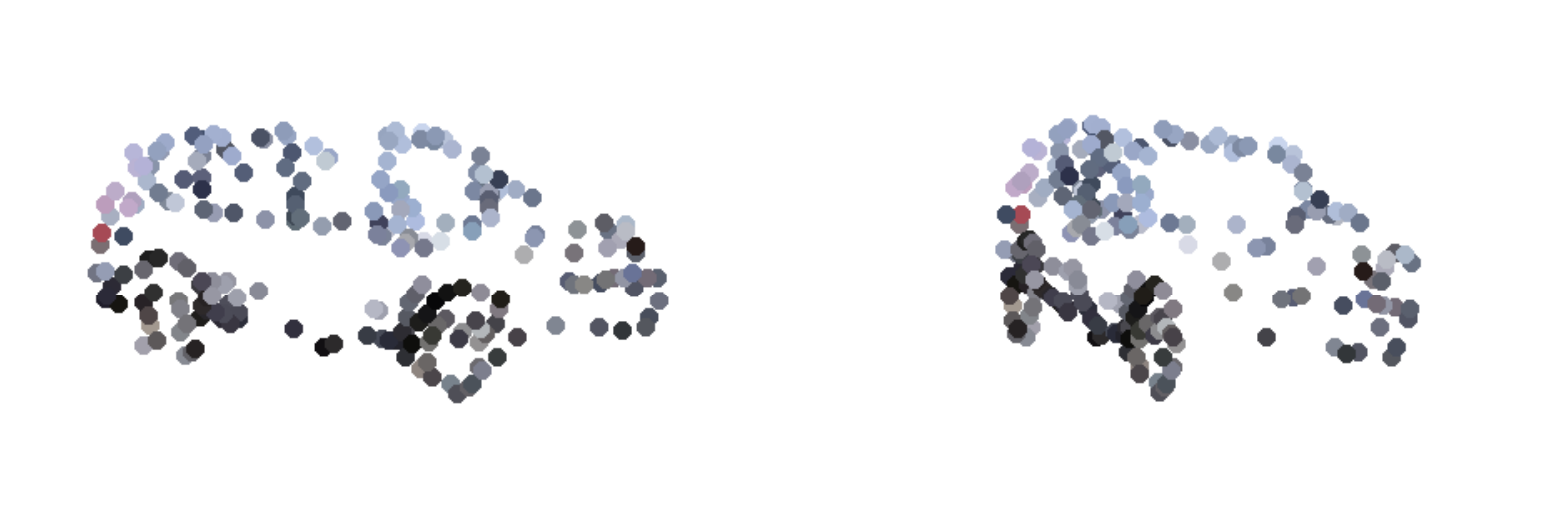} \\[-2Ex]
\caption{Matching different cars. Best viewed in color. Left: the correspondences of sedan images and the reconstruction. Right: the correspondences of SUV images and the reconstruction. Only four selected images are shown for each image set. Note that the cars in images are all different and the feature points are automatically detected. The markers with the same color indicate the matched feature points. The 3D reconstruction is rendered with the colors in the first image and visualized in two viewpoints.} \label{fig:cars}
\end{figure*}

Recent years have witnessed growing interest in reconstructing category-specific object models from single images, which is still an open problem \cite{vicente2014reconstructing,carreira2014virtual}. Among a series of challenges, feature matching for different object instances is the foremost and previous work usually assumed that correspondences of some keypoints were given \cite{vicente2014reconstructing,carreira2014virtual}. In this section, we demonstrate the applicability of joint matching to solve this problem.

We use the FG3DCar datasets \cite{Lin2014jointly} and try to match the images of different car models in the same category (e.g., sedan or SUV). Following the general practice in object reconstruction \cite{vicente2014reconstructing,carreira2014virtual}, we assume segmentation is provided such that background can be ignored, and we only match images with similar views. We select nine sedans and eight SUVs and match two sets of images separately. Note that the car models are all different from each other. See \refFig{fig:cars} for examples.

To exact descriptive features we first detect image edges by the structured forests \cite{dollar2013structured} and sample a number of points on the edges with constant spacing. On average, we obtain $\sim$600 feature points for each image. Since the object appearance is changed from image to image and the features are automatically extracted, the matching is extremely difficult. Inspired by recent works \cite{weinzaepfel2013deepflow,carreira2014virtual}, we adopt deep features, i.e., middle-layer responses of convolutional neural nets (CNN), as descriptors for feature matching. More specifically, we use the publicly available deep learning toolbox Caffe \cite{jia2014caffe} and the pre-trained CNN Alexnet \cite{krizhevsky2012imagenet}. We feed a $192\times 192$ patch around each feature point forward through the Alexnet. The center columns of conv4 and conv5 layers are concatenated and normalized to form a 640 dimensional feature vector. To leverage the prior on object rigidity, we use pairwise graph matching solved by the Reweighted Random Walk algorithm \cite{cho2010reweighted} and collect the output scores of candidate matches as affinity scores. Then, we delete the points with candidate matches in less than two images and run MatchALS.

\begin{figure}
\centering
\includegraphics[width=0.5\linewidth]{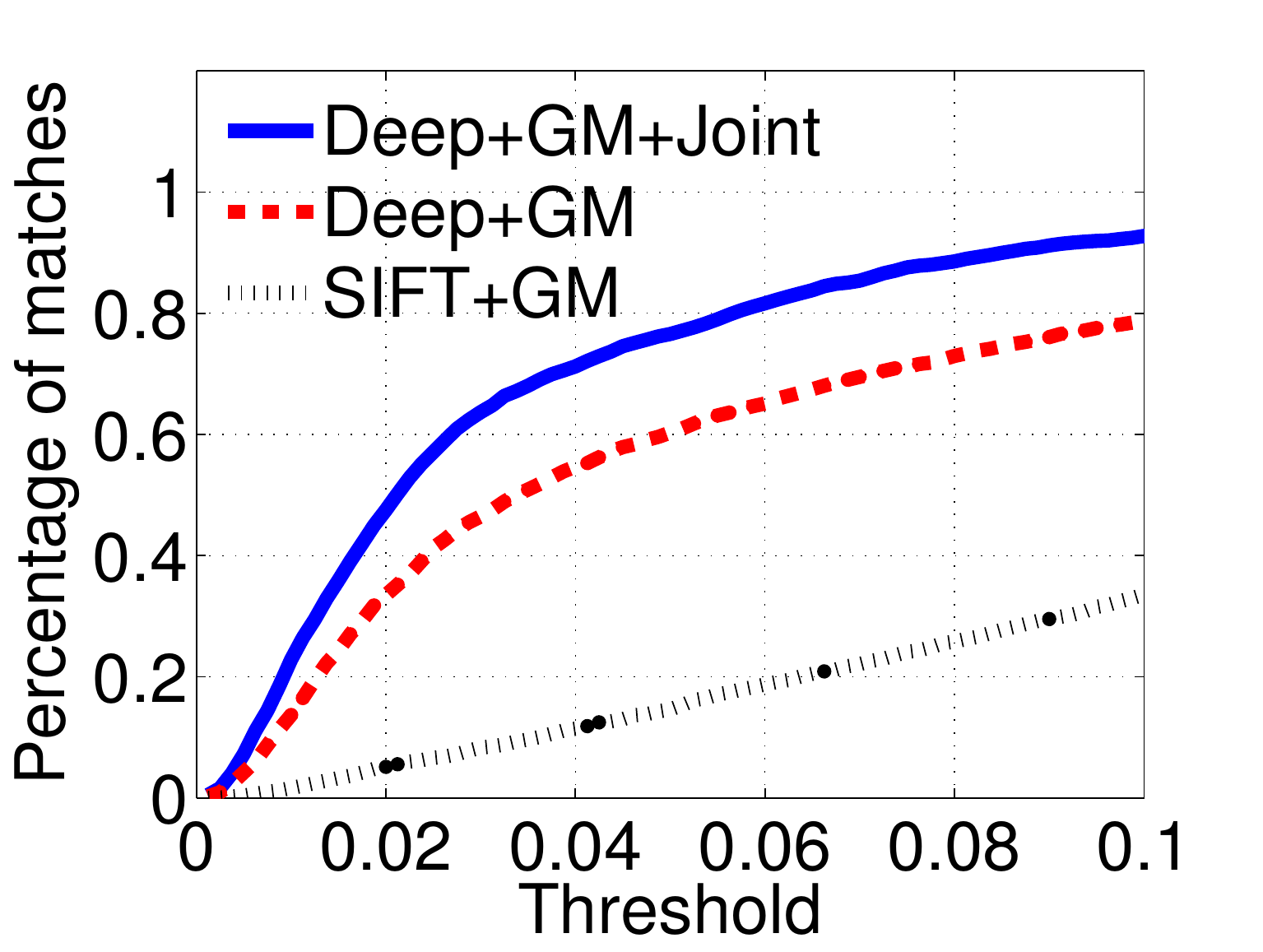}
\caption{The performance curve of car image matching. ``Deep" represents deep features. ``GM" denotes graph matching. ``Joint" means joint matching using the proposed method.} \label{fig:cars-plot}
\vspace{-1em}
\end{figure}

We adopt the same metric introduced in \refSec{sec:graffiti} for quantitative evaluation and use the manually-annotated landmarks provided in the datasets as ground truth. The result on the sedan images is shown in \refFig{fig:cars-plot}. Matching with SIFT features fails since local image patterns are different for two cars. Graph matching with deep features obtains a much better performance, which is further improved by the proposed joint matching algorithm. We obtain a very similar result on the SUV images, which is not plotted.

The results are visualized in \refFig{fig:cars}. The corresponding parts of cars are basically matched in spite of the large differences in appearances and viewpoints. Note that the features are automatically detected and therefore not fully overlapped for two images. For a simple demonstration, we run rigid reconstruction from the estimated feature correspondences by triangulation with an orthographic camera model and the viewpoints provided in the dataset. Despite some noises and missing points, we can clearly see the 3D structures of a sedan and a SUV. We believe that more appealing reconstructions can be obtained by using sophisticated reconstruction techniques and more information such as object silhouette and surface smoothness, while they are out of the scope of this paper.

\section{Conclusion}

In this paper, we proposed a practical solution to multi-image matching. We use pairwise feature similarities or graph matching scores as input and obtain accurate matches by an efficient algorithm that globally optimizes for both feature affinities and cycle consistency of matches. The experiments not only validate the effectiveness of the proposed method but also demonstrate that joint matching is a promising approach to matching images with different object instances as the first step towards reconstructing object models from crowd-sourced image collections. As future work, we would like to explore more applications and incremental algorithms for joint matching.

\bigskip

\small
\noindent\textbf{Acknowledgments}: Grateful for support through the following grants:
NSF-DGE-0966142,
NSF-IIS-1317788,
NSF-IIP-1439681,
NSF-IIS-1426840,
ARL RCTA W911NF-10-2-0016, and
ONR N000141310778

\footnotesize
\bibliographystyle{ieee}
\bibliography{mybib-abbr}

\end{document}